\documentclass{article}

\usepackage[preprint]{neurips_2022}

\usepackage[utf8]{inputenc} 
\usepackage[T1]{fontenc}    
\usepackage{hyperref}       
\usepackage{xurl}
\usepackage{booktabs}       
\usepackage{amsfonts}       
\usepackage{nicefrac}       
\usepackage{microtype}      
\usepackage{xcolor}         

\usepackage{amsmath}
\usepackage{mathtools}
\usepackage{amssymb}
\usepackage{pifont}
\usepackage{times}
\usepackage{latexsym}
\usepackage{booktabs}
\usepackage{multirow}
\usepackage{adjustbox}
\usepackage{siunitx}
\usepackage{graphicx}

\hypersetup{
           breaklinks=false,   
           colorlinks=true,   
           urlcolor=blue
        }

\usepackage{array}
\newcolumntype{P}[1]{>{\centering\arraybackslash}p{#1}}
\newcolumntype{L}[1]{>{\raggedright\arraybackslash}p{#1}}

\newcolumntype{F}[1]{%
    >{\raggedright\arraybackslash\hspace{0pt}}p{#1}}%
\newcolumntype{T}[1]{%
    >{\centering\arraybackslash\hspace{0pt}}p{#1}}%

\title{On the Effectiveness of Compact Biomedical Transformers}

\author{%
  Omid Rohanian\textsuperscript{1,$\dagger$} \\
  \texttt{omid.rohanian@eng.ox.ac.uk} \\
   \And
   Mohammadmahdi Nouriborji\textsuperscript{4,$\dagger$} \\
   \texttt{m.nouriborji@nlpie.com} \\
   \AND
   Samaneh Kouchaki\textsuperscript{2} \\
   \texttt{samaneh.kouchaki@surrey.ac.uk} \\
   \And
   David A. Clifton\textsuperscript{1,3} \\
   \texttt{david.clifton@eng.ox.ac.uk} \\
}

\begin{document}

\maketitle

\def\thefootnote{$\dagger$}\footnotetext{The two authors contributed equally to this work.}
\def\thefootnote{\arabic{footnote}}

\begin{center}

\textsuperscript{1}Department of Engineering Science, University of Oxford, Oxford, UK

\textsuperscript{2}Dept. Electrical and Electronic Engineering, University of Surrey, Guildford, UK

\textsuperscript{3}Oxford-Suzhou Centre for Advanced Research, Suzhou, China

\textsuperscript{4}NLPie Research, Oxford, UK 

\end{center}

\begin{abstract}
  Language models pre-trained on biomedical corpora, such as BioBERT, have recently shown promising results on downstream biomedical tasks. Many existing pre-trained models, on the other hand, are resource-intensive and computationally heavy owing to factors such as embedding size, hidden dimension, and number of layers. The natural language processing (NLP) community has developed numerous strategies to compress these models utilising techniques such as pruning, quantisation, and knowledge distillation, resulting in models that are considerably faster, smaller, and subsequently easier to use in practice. By the same token, in this paper we introduce six lightweight models, namely, BioDistilBERT, BioTinyBERT, BioMobileBERT, DistilBioBERT, TinyBioBERT, and CompactBioBERT which are obtained either by knowledge distillation from a biomedical teacher or continual learning on the Pubmed dataset via the Masked Language Modelling (MLM) objective. We evaluate all of our models on three biomedical tasks and compare them with BioBERT-v1.1 to create efficient lightweight models that perform on par with their larger counterparts. All the models will be publicly available on our Huggingface profile at \url{https://huggingface.co/nlpie} and the codes used to run the experiments will be available \href{https://github.com/nlpie-research/Compact-Biomedical-Transformers}{on our Github page}.  

\end{abstract}

\section{Introduction}
\label{intro}

\begin{figure*}[ht!]
\centering
\includegraphics[scale=0.25]{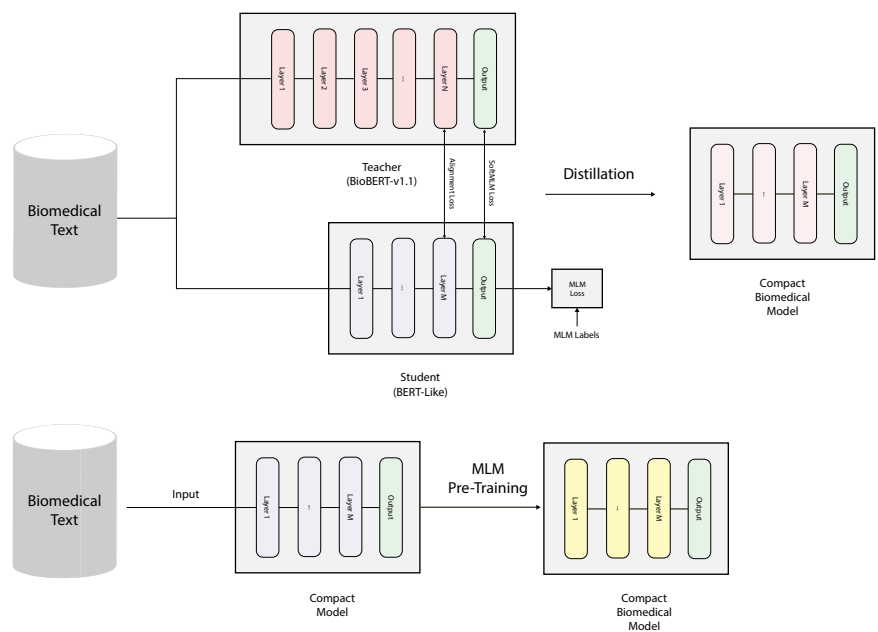}
\caption{The two general strategies proposed for training compact biomedical models. The first approach is to directly distil a compact model from a biomedical teacher which in our work is BioBERT-v1.1. The distillation depicted in this figure is the same technique used for obtaining DistilBioBERT. TinyBioBERT and CompactBioBERT, on the other hand, employ different approaches, which are not shown here. The second method involves additionally pre-training a compact model on biomedical corpora. For this approach, we use compact models which have been distilled from powerful teachers, namely, DistilBERT \citep{sanh2019distilbert}, TinyBERT \citep{jiao-etal-2020-tinybert}, and MobileBERT \citep{sun-etal-2020-mobilebert}.}
\label{fig:architecture}
\end{figure*}

There has been an ever-increasing abundance of medical texts in recent years, both in private and public domains, which provide researchers with the opportunity to automatically process and extract useful information to help develop better diagnostic and analytic tools \citep{locke2021natural}. Medical corpora can come in various forms, each with its own specific context. These include Electronic Health Records (EHR), medical texts on social media, online knowledge bases, and scientific literature \citep{KALYAN2020103323}. 

Recent advances in Natural Language Processing (NLP) and deep learning have made it possible to computationally process biomedical texts as varied as the above using powerful generic methods that learn text representations based on contextual information around each word. These methods alleviate the need for cumbersome feature engineering or extensive preprocessing, and when combined with the appropriate GPU technology, can handle large volumes of data with a high level of efficiency \citep{wu2020deep}. Contextualised embeddings like ELMo \citep{peters-etal-2018-deep} and BERT \citep{devlin-etal-2019-bert}, while having been derived primarily using a generic language modelling objective, are able to capture task-agnostic and generalisable syntactic and semantic properties of words in their context, making them useful for various downstream applications \citep{ethayarajh2019contextual, tenney2019you}.    

In recent years, different `probing' methods have been developed to study different aspects of word embeddings and understand their internal mechanics \citep{conneau-etal-2018-cram,jawahar2019does, clark2019does}. These studies have shown that BERT encapsulates a surprising amount of knowledge about the world and can be used to solve tasks that traditionally would require encoded information from knowledge-bases \citep{rogers2020primer}. These models are not without their own drawbacks and come with certain limitations. For example, it has been shown that BERT does not understand negation by default \citep{ettinger2020bert} or struggles with representations of numbers \citep{wallace2019nlp}. Regardless of these shortcomings, BERT and its different variants are still the state-of-the-art in different areas of NLP. 

With the advent of the transformers architecture \citep{vaswani2017attention}, the NLP community has moved towards utilising pre-trained models that could be used as a strong baseline for different tasks and also serve as a backbone to other sophisticated models. The standard procedure is to use a general model pre-trained on a very large amount of unstructured text and then fine-tune the model and adapt it to the specific characteristics of each task. Most state-of-the-art NLP models are based on this procedure.   

A related alternative to the standard pretrain and fine-tune approach is domain-adaptive pretraining, which has been shown to be effective on different textual domains. In this paradigm, instead of fine-tuning the pretrained model on the task-specific labelled data, pretraining continues on the unlabeled training set. This allows a smaller pretraining corpus, but one that is assumed to be more relevant to the final task \citep{gururangan-etal-2020-dont}. This method is also known as continual learning, which refers to the idea of incrementally training models on new streams of data while retaining prior knowledge \citep{PARISI201954}.  

NLP researchers working with biomedical data have naturally started to incorporate these techniques into their models. Apart from vanilla fine-tuning on medical texts, specialised BERT-based models have also been developed that are specifically trained on medical and clinical corpora. ClinicalBERT \citep{huang2019clinicalbert}, SciBERT \citep{beltagy2019scibert}, and BioBERT \citep{lee2020biobert} are successful attempts at developing pretrained models that would be relevant to biomedical NLP tasks. They are regularly used in the literature to develop the latest best performing models on a wide range of tasks. 

Regardless of the successes of these architectures, their applicability is limited because of the large number of parameters they have and the amount of resources required to employ them in a real setting. For this reason, there is a separate line of research in the literature to create compressed versions of larger pretrained models with minimal performance loss. DistilBERT \citep{sanh2019distilbert}, MobileBERT \citep{sun-etal-2020-mobilebert}, and TinyBERT \citep{jiao-etal-2020-tinybert} are prominent examples of such attempts, which aim to produce a lightweight version of BERT that closely mimics its performance while having significantly less trainable parameters. The process used in creating such models is called distillation \citep{hinton2015distilling}.


In this work we first train three distilled versions of the BioBERT-v1.1 using different distillation techniques, namely, DistilBioBERT, CompactBioBERT, and TinyBioBERT. Following that, we pre-train three well-known compact models (DistilBERT, TinyBERT, and MobileBERT) on the PubMed dataset using continual learning. The resultant models are called BioDistilBERT, BioTinyBERT, and BioMobileBERT. Finally, we compare our models to BioBERT-v1.1 through a series of extensive experiments on a diverse set of biomedical datasets and tasks. The analyses show that our models are efficient compressed models that can be trained significantly faster and with far fewer parameters compared to their larger counterparts, with minimal performance drops on different biomedical tasks. To the best of our knowledge, this is the first attempt to specifically focus on training compact models on biomedical corpora and by making the models publicly available we provide the community with a resource to implement powerful specialised models in an accessible fashion.     

The contributions of this paper can be summarised as follows:

\begin{itemize}
    \item We are the first to specifically focus on training compact biomedical models using distillation and continual learning.
    \item Utilising continual learning via the Masked Language Modelling (MLM) objective, we train three well-known pre-trained compact models, namely DistilBERT, MobileBERT, and TinyBERT for $200$k steps on the PubMed dataset.
    \item We distil three students from a biomedical teacher (BioBERT-v1.1) using three different distillation procedures, which generated the following models: DistilBioBERT, TinyBioBERT, and CompactBioBERT. 
    \item We evaluate our models on a wide range of biomedical NLP tasks that include Named Entity Recognition (NER), Question Answering (QA), and Relation Extraction (RE).    
    \item We make all of our $6$ compact models freely available on Huggingface and Github. These models cover a wide range of parameter sizes, from $15$M parameters for the smallest model to $65$M for the largest. 
\end{itemize}


\section{Background}

Pre-training followed by fine-tuning has become a standard procedure in many areas of NLP and forms the backbone for most state-of-the-art models such as BERT \citep{devlin-etal-2019-bert} and GPT-3 \citep{brown2020language}. The goal of language model pre-training is to acquire effective in-context representations of words based on a large corpus of text, such as Wikipedia. This process is often self-supervised, which means that the representations are learned without using human-provided labels. There are two main strategies for self-supervised pre-training, namely, MLM and Causal Language Modeling (CLM). In this work, we focus on models pre-trained with the MLM objective. 

\subsection{Masked Language Modeling}
MLM is the process of randomly omitting portions of a given text and having the model predict the omitted portions. The masking percentage is normally $15\%$, with an $80\%$ probability that the selected word will be substituted with a specific mask token (e.g. <MASK>) and a $20\%$ chance that it will be replaced with another random word \citep{devlin-etal-2019-bert}. Contextualised representations generated using these pre-trained language models are referred to as bidirectional, which means that information from previous and following contexts is used to construct representations for each given word.

MLM utilises distributional hypothesis, an idea introduced originally by \citet{harris1954distributional} and later popularised by \citet{firth1957synopsis}. The premise is that words that occur in the same contexts tend to have a similar meaning, or as Firth phrased it, ``a word is characterised by the company it keeps''.  As a result, BERT shares conceptual similarities with other representation learning schemes in NLP. There is strong evidence to suggest that MLM relies on distributional semantic information significantly more than grammatical structure of sentences \citep{sinha-etal-2021-masked}.   

\subsection{BERT: Bidirectional Encoder Representation from Transformers}
The most prominent transformer pre-trained with MLM is BERT. BERT is an encoder-only transformer that relies on the Multi-Head Attention mechanism for learning in-context representations. BERT has different variations such as $BERT_{base}$ and $BERT_{large}$ which vary in the number of layers and the size of the hidden dimension. Original BERT is trained on English Wikipedia and BooksCorpus datasets for about 1 million training steps, making it a strong model for various downstream NLP tasks. 

Fine-tuning pre-trained BERT on a downstream task involves training the model for a few more epochs using a labelled dataset and with a lower learning rate  \citep{sun2019fine}. It has been shown that, since this procedure only affects the weights in the top layers of BERT, it will not lead to catastrophic forgetting of linguistic information \citep{merchant2020happens}. 

\subsection{BioBERT and other Biomedical Models}

While generic pre-trained language models can perform reasonably well on a variety of downstream tasks even in domains other than those on which they have been trained, in recent years researchers have shown that continual learning and pre-training of language models on domain-specific corpora leads to noticeable performance boosts compared to simple fine-tuning. BioBERT is an example of such a domain-specific BERT-based model and the first that is trained on biomedical corpora. 

BioBERT takes its initial weights from $BERT_{base}$ (pre-trained on Wikipedia + Books) and is further pre-trained using the MLM objective on the PubMed and optionally PMC datasets. BioBERT has shown promising performance in many biomedical tasks including NER, RE, and QA. Aside from BioBERT, numerous additional models have been trained entirely or partially on biomedical data, including ClinicalBERT \citep{huang2019clinicalbert}, SciBERT \citep{beltagy-etal-2019-scibert}, BioMedRoBERTa \citep{gururangan-etal-2020-dont}, and BioELECTRA \citep{raj2021bioelectra}. 

\subsection{Overparametrisation of Language Models}

The $BERT_{base}$ model has $110$M parameters, which is a modest number compared to T5 ($111$B), GPT-3 ($175$B), or MT-NLG ($530$B). Training models of this magnitude comes with considerable financial and environmental costs. This trend is unlikely to be reversed anytime soon given the increasing computational power and the resources that large technology companies devote to creating such models \citep{bender2021dangers}. 

\citet{strubell2019energy} studied several major transformer-based models and estimated the carbon footprint and cloud compute costs incurred during their training.  Warning against environmentally unfriendly practices in AI and NLP research has created interest in the community to develop lighter but computationally efficient models that come with minimal reduction in performance. This trend has been described as `Green AI' \citep{schwartz2020green}. Model compression can be considered a step in this direction. It is predicated on the idea of creating a quick and compact model to imitate a slower, bigger, but more performant model \citep{bucilua2006model}. Several different model compression methods exist, with the aim to encode large models and create smaller more compact versions of them. The present work focuses on knowledge distillation but we will also briefly mention quantisation and pruning. 

\subsection{Quantisation and Pruning}

Quantisation is a technique that attempts to reduce the memory footprint of a pre-trained language model by reducing the precision of its weights and uses low bit hardware operations to speed up computation \citep{shen2020q}. It is an effective method for model compression and acceleration that can be applied to both pre-trained models or models trained from scratch \citep{cheng2017survey}. This method requires hardware compatibility to function \citep{rogers2020primer}. 

Pruning is another model compression method that disables certain parts of a larger model to create a compressed faster version of it. It has been shown that zeroing out different parts of the multi-head attention mechanism in BERT does not result in a significant drop during inference time \citep{michel2019sixteen}. Pruning can be performed in a structured way, where certain components of the model are removed, or in an unstructured fashion, where weights are dropped regardless of location in the network \citep{rogers2020primer}. Since quantisation and pruning are independently developed and complementary to each other, they can be used in tandem to develop a single compressed model. 

\subsection{Knowledge Distillation}

Knowledge distillation \citep{hinton2015distilling} is the process of transferring knowledge from a larger model called ``teacher'' to a smaller one called ``student'' using the larger model's outputs as soft labels. Distillation can be done in a task-specific way where the pre-trained model is first fine-tuned on a task and then the student attempts to imitate the teacher network. This is an effective method, however, fine-tuning of a pre-trained model can be computationally expensive. Task-agnostic distillation, on the other hand, allows the student to mimic the teacher by looking at its masked language predictions or intermediate representations. The student can subsequently be directly fine-tuned on the final task \citep{wang2020minilm, yao2021adapt}. 

DistilBERT is a well-known example of a compressed model that uses knowledge distillation to transfer the knowledge within the $BERT_{base}$ model to a much smaller student network which is about $40$\% smaller and $60$\% faster. It uses a triple loss which is a linear combination of language modeling, distillation and cosine-distance losses.



\section{Approach}

In this work, we focus on training compact transformers on biomedical corpora. Among the available compact models in the literature, we use DistilBERT, MobileBERT, and TinyBERT models which have shown promising results in NLP. We train compact models using two different techniques as shown in Figure \ref{fig:architecture}. The first is continual learning of pre-trained compact models on biomedical corpora. In this strategy, each  model is further pre-trained on the PubMed dataset for $200$k steps via the MLM objective. The obtained models are named BioDistilBERT, BioMobileBERT, and BioTinyBERT.

For the second strategy, we employ three distinct techniques: the DistilBERT and TinyBERT distillation processes, as well as a mixture of the two. The obtained models  are named DistilBioBERT, TinyBioBERT, and CompactBioBERT. We test our models on three well-known biomedical tasks and compare them with BioBERT-v1.1 as shown in Tables \ref{t:distilledNER} to \ref{qa:continual}.

\section{Methods}

\begin{figure*}[ht!]
\centering
\label{fig:efficieny}
\includegraphics[scale=0.35]{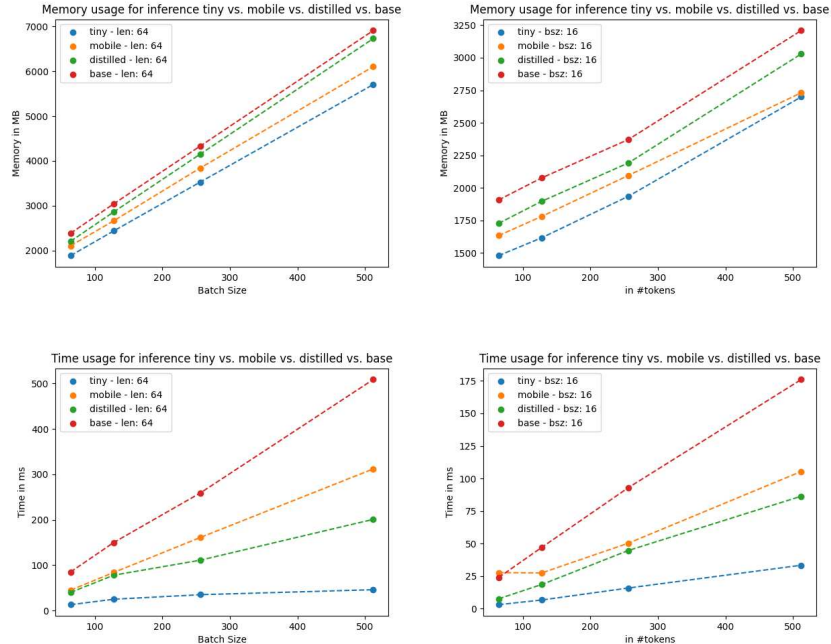}
\caption{The inference time/memory comparison of our proposed models. `small' refers to TinyBioBERT, `mobile' to MobileBioBERT, `distilled' to DistilBioBERT and CompactBioBERT (since they share the same architecture), and `base' to BioBERT-v1.1.}
\end{figure*}

In this section, we describe the internal architecture of each compact model that is explored in the paper, the method used to initialise its weights, and the distillation procedure employed to train it.  


\subsection{DistilBioBERT}
\label{distilbiobert}

\subsubsection{Architecture}
In this model, the size of the hidden dimension and the embedding layer are both set to $768$. The vocabulary size is $28996$ for the cased version which is the one employed in our experiments. The number of transformer layers is $6$ and the expansion rate of the feed-forward layer is $4$. Overall this model has around $65$ million parameters.

\subsubsection{Initialisation of the Student}
Effective initialisation of the student model is critical due to the size of the model and the computational cost of distillation.
As a result, there are numerous techniques available for initialising the student. One method introduced by \citet{distillation_google} is to initialise the student via MLM pre-training and then perform distillation. Another approach, which we have followed in this work, is to take a subset of the larger model by using the same embedding weights and initialising the student from the teacher by taking weights from every other layer \citep{sanh2019distilbert}. With this approach, the hidden dimension of the student is restricted to that of the teacher model.  

\subsubsection{Distillation Procedure}
For distillation, we mainly follow the work of \citet{sanh2019distilbert} in which the loss is a combination of three different terms. In this section, we explain each of these in detail. The first term is normal cross entropy loss used for the MLM objective which can be expressed with the below equation:

\begin{equation} \label{eq:1}
    L_{mlm}(X , Y) = -\sum^{N}_{n=1}W_{n}(\sum^{|V|}_{i=1} Y^{n}_{i}ln(f_{s}(X)_{i}^{n}))
\end{equation}
where $X$ is the input of the model, $Y$ denotes MLM labels which is a collection of $N$ one-hot vectors each with the size of $|V|$ where $|V|$ is the size of the vocabulary of the model and $N$ is the number of input tokens\footnote{Note that one-hot vectors for non-masked tokens are zero vectors.} and $W_{n}$ is $1$ for masked tokens and $0$ for others. This ensures that only masked tokens will contribute to the computation of loss. $f_{s}$ represents the student model whose output is a probability distribution vector with the size of the vocabulary ($|V|$) for each token. 

The second loss term used for distillation is a KL Divergence loss over the outputs (aka soft labels) of the teacher model which can be expressed in the below equation where $f_{t}$ represents the teacher model:


\begin{equation} \label{eq:2}
    L_{softMLM}(X) = -\sum^{N}_{n=1}W_{n}D_{KL}(f_{t}(X )^{n}_{i} \hspace{5pt} || \hspace{5pt} f_{s}(X)_{i}^{n})
\end{equation}

Finally, there is an optional loss that is intended to align the last hidden state of the teacher and student models via a cosine embedding loss:

\begin{equation}  \label{eq:3}
    L_{align}(X) = \frac{1}{N} \sum^{N}_{n=1} 1 - \phi(h_{t}(X )^{n}, h_{s}(X)^{n})
\end{equation}
where $h_{t}$ and $h_{s}$ represent functions that output the last hidden state of the teacher and student models respectively (each of which is a collection of $N$, $D$-dimensional vectors where $D$ is the size of the hidden dimension) and $\phi$ is a cosine similarity function\footnote{Cosine similarity is expressed with the formula: $\phi(\Vec{u} , \Vec{v}) = \frac{\Vec{u}.\Vec{v}}{||\Vec{u}||_{2}||\Vec{v}||_{2}}$}.

Finally, the combined distillation loss can be expressed as follows:

\begin{align} 
    L(X , Y) = &\hspace{12pt}\alpha_{1}L_{mlm}(X , Y) \\
               & + \alpha_{2}L_{softMLM}(X) \nonumber \\ 
               & + \alpha_{3}L_{align}(X) \nonumber
\end{align}
where $\alpha_1$, $\alpha_2$ and $\alpha_3$ are weighting terms for combining different losses. In our settings $\alpha_1 = 2.0$, $\alpha_2 = 5.0$, and $\alpha_3 = 1.0$.

\subsection{TinyBioBERT}
This model uses a unique distillation method called `transformer-layer distillation' which is applied on each layer of the student to align the attention maps and the hidden states of the student with the teacher.

\subsubsection{Architecture}
This model is available in two sizes: The first one is a $4$-layer transformer with a hidden dimension and embedding size of $312$ and about $15$M parameters.
The second is a $6$-layer transformer with the same design as DistilBERT, as  described in Section \ref{distilbiobert}.
This model contains around $30.5$K words in its vocabulary and employs an uncased tokeniser, which means it does not include upper-cased letters in its vocabulary. 

\subsubsection{Initialisation of the Student}
The initial weight initialisation of this model is random since the hidden and the embedding size of this model differ from its teacher. However, the weight initialisation of the DistilBERT can be used when the hidden and embedding size of the student are the same as the ones in the teacher which to the best of our knowledge was not tried in the original paper.

\subsubsection{Transformer-layer distillation}
This distillation is applied on attention maps and outputs of each transformer layer of the student along with the final output layer and embedding layer of the student. Since the student is smaller than the teacher, the numbers of layers are not equal. As a result, each layer of the student will be mapped to a specific layer of the teacher with which the distillation will be performed. The mapping from the student layer index to the corresponding teacher layer index is determined by the equation below:

\begin{equation}
    T_i = g(i)
\end{equation}
where $i$ is the index of the student layer, $g(.)$ is the mapping function, and $T_i$ is the index of the respective transformer layer of the teacher. In both models, $g(0) = 0$ which is the index of the embedding layer and $g(M + 1) = N + 1$ which is the index of the output layer.

The mean squared error loss between each student layer and its corresponding layer in the teacher is calculated as follows: 

\begin{align} \label{eq:7}
    L_{Layer}(X, l) =  & \hspace{5pt} MSE(h_s^l(X) W_{h} , h_t^{g(l)}(X))\\
                     & + \frac{1}{H}\sum^H_{i=1} MSE(a_s^l(X)^i, a_t^{g(l)}(X)^i) \nonumber
\end{align}
where $h_s^l(X)$ and $h_t^{g(l)}(X)$ will output the hidden states of the $l_{th}$ layer of the student and the $g(l)_{th}$ of the teacher respectively. $a_s^l(X)$ and $a_t^{g(l)}(X)$ will output the attention maps of the $l_{th}$ layer of the student and the $g(l)_{th}$ of the teacher, respectively. Because these models use multi-head attention, we have $H$ attention maps per layer, and the mean squared error is applied to each head independently, as shown in the Equation \ref{eq:7}. Finally, $W_{h}$ is a projection weight used when the hidden dimensions of the student and the teacher are not the same.

In addition to the transformer-layer loss described above, TinyBERT use two additional losses, one for the embedding layer and one for the student's output probabilities. The embedding loss is designed to align the embedding of the student ($E_s$) with that of the teacher ($E_t$). This loss is only required if the student and teacher do not share the same embedding layer. The embedding loss is expressed in the below equation:

\begin{equation}
    L_{Embed} = MSE(E_{s} W_{e} , E_t)
\end{equation}
where $W_{e}$ is a projection weight as discussed in Equation \ref{eq:7}. TinyBERT employs one additional loss to align the final probability distributions of teacher and student, which is a cross entropy loss over the teacher's soft labels:

\begin{equation}\label{eq:8}
    L_{output}(X) = -\frac{1}{N}\sum^{N}_{n=1}\sum^{|V|}_{i=1} f_{t}(X)^{n}_{i}ln(f_{s}(X)_{i}^{n})
\end{equation}

The complete loss function used for TinyBERT distillation is as follows:

\begin{align}
    L(X) = & \hspace{12pt} \lambda_{0}L_{Embed} \\
           & + \sum_{l=1}^{M} \lambda_{l}L_{Layer}(X, l) \nonumber \\
           & + \lambda_{(M + 1)}L_{output}(X) \nonumber
\end{align}
where $\lambda_0$ to $\lambda_{(M + 1)}$ are hyperparameters, controlling the importance of each layer. In this work all lambdas are set to $1.0$.

\subsection{CompactBioBERT}
\label{compactbiobert} 

This model has the same overall architecture as DistilBioBERT (Section \ref{distilbiobert}), with the difference that here we combine the distillation approaches of DistilBERT and TinyBERT. We utilise the same initialisation technique as in DistilBioBERT, and apply a layer-to-layer distillation with three major components, namely, MLM, layer, and output distillation.

Layer distillation is performed between each student layer and its corresponding teacher layer based on Equation \ref{eq:7}, with the MSE losses substituted with cosine embedding loss for hidden states alignment and KL Divergence for attention maps alignment. Below is the final layer distillation loss proposed for CompactBioBERT:

\begin{align} 
    L_{compact}(X, l) =  & \hspace{5pt} \frac{1}{N} \sum_{n=1}^N 1 - \phi(h_s^l(X)^n , h_t^{g(l)}(X)^n)\\
                     & + \frac{1}{HN}\sum^H_{i=1} \sum^{N}_{n=1} D_{KL}(a_s^l(X)^i_n \hspace{5pt} || \hspace{5pt} a_t^{g(l)}(X)^i_n) \nonumber
\end{align}

The MLM and output distillations are the same losses used in DistilBioBERT. MLM distillation corresponds to $L_{mlm}(X, Y)$ in Equation \ref{eq:1} and $L_{softMLM}(X)$ denotes output distillation from Equation \ref{eq:2}. Finally, the complete distillation loss used in CompactBioBERT is as follows:

\begin{align}
    L(X , Y) = &\hspace{12pt}\alpha_{1}L_{mlm}(X , Y) \\
               & + \alpha_{2}L_{softMLM}(X) \nonumber \\ 
               & + \alpha_{3}\sum_{l=1}^M L_{compact}(X,l) \nonumber
\end{align}
where $\alpha_1$, $\alpha_2$, and $\alpha_3$ are weighting terms for combining different losses. In our settings, $\alpha_1 = 1.0$, $\alpha_2 = 5.0$, and $\alpha_3 = 3.0$.

\subsection{BioMobileBERT}
MobileBERT \citep{sun-etal-2020-mobilebert} is a compact model that uses a unique design comprised of different components to reduce the model's width (hidden size) while maintaining the same depth as $BERT_{large}$ ($24$ Transformer Layers). MobileBERT has proved to be competitive in many NLP tasks while also being efficient in terms of both computational and parameter complexity.

\subsubsection{Architecture and Initialisation}
MobileBERT uses a $128$-dimensional embedding layer followed by $1$D convolutions to up-project its output to the desired hidden dimension expected by the transformer blocks. For each of these blocks, MobileBERT uses linear down-projection at the beginning of the transformer block and up-projection at its end, followed by a residual connection originating from the input of the block before down-projection. Because of these linear projections, MobileBERT can reduce the hidden size and hence the computational cost of multi-head attention and feed-forward blocks. This model additionally incorporates up to four feed-forward blocks in order to enhance its representation learning capabilities. Thanks to the strategically placed linear projections, a $24$-layer MobileBERT (which is used in this work) has around $25$M parameters. To the best of our knowledge MobileBERT is initialised from scratch.

\subsubsection{Distillation Procedure}
MobileBERT uses layer-wise distillation similar to TinyBERT \citep{jiao-etal-2020-tinybert} and CompactBioBERT (Sec. \ref{compactbiobert}). Unlike TinyBERT, where the student's hidden dimension and number of layers may differ from those of the teacher, MobileBERT utilises a unique teacher named IB-BERT which has the same hidden dimension and number of layers as the student \footnote{Since MobileBERT's teacher is a custom variant of the $BERT_{large}$ called IB-BERT, we were not able to distil a compact model with the same procedure as MobileBERT. Therefore, we solely pre-trained MobileBERT on the PubMed dataset via MLM objective and continual learning.}. As a result, mapping each transformer layer in the student to its matching teacher layer is unnecessary.

The loss employed by MobileBERT for layer-wise distillation is shown below: 

\begin{align} \label{eq:12}
    L_{mobile}(X, l) =  & \hspace{5pt} MSE(h_s^l(X) , h_t^{l}(X))\\
                        & + \frac{1}{H}\sum^H_{i=1} \sum^{N}_{n=1} D_{KL}(a_s^l(X)^i_n \hspace{5pt} || \hspace{5pt} a_t^{l}(X)^i_n) \nonumber \nonumber
\end{align}

The {loss}\footnote{Note that the original formula contains a Next Sentence Prediction (NSP) loss term as well which is omitted here for brevity.} used for distillation of the MobileBERT is as follows:

\begin{align}
    L(X , Y) = &\hspace{5pt}\alpha L_{mlm}(X , Y) \\
               & + (1 - \alpha)(\frac{1}{M}\sum_{l=1}^M L_{mobile}(X,l)) \nonumber
\end{align}
where $M$ is the number of transformer layers and  $\alpha$ is a hyperparameter between $(0 , 1)$.

\section{Experiments and Results}

We evaluate our models on three biomedical tasks, namely, NER, QE, and RE. For a fair comparison, we fine-tune all of our models using a constant seed. Note that the results obtained in this work are for comparison with BioBERT-v1.1 in a similar setting and we are not focusing on reproducing or outperforming state-of-the-art on any of the datasets since that is not the objective of this work. 

We distil our students solely from BioBERT and also compare our continually learnt models with it. While there are other recent biomedical transformers available in the literature (Sec. \ref{intro}), BioBERT is the most general (trained on large biomedical corpora for $1$M steps) and is widely used as a backbone for building new architectures. Direct comparison with one major model helps us to keep the work focused on compression techniques and assessing their efficiency in preserving information from a well-performing and reliable teacher. These experiments can in the future be expanded to cover other biomedical models.    

For biomedical NER we use 8 well-known datasets, namely, NCBI-disease \citep{dougan2014ncbi}, BC5CDR (disease and chem) \citep{li2016biocreative}, BC4CHEMD \citep{krallinger2015chemdner}, BC2GM \citep{smith2008overview}, JNLPBA \citep{kim2004introduction}, LINNAEUS \citep{gerner2010linnaeus}, and Species-800 \citep{pafilis2013species} which will test the biomedical knowledge of our models in different categories such as Disease, Drug/chem, Gene/protein, and Species. All of our models were trained for $5$ epochs with a batch size of $16$ and a learning rate of $5e-5$. In a few cases, a learning rate of $3e-5$ and a batch size of $32$ were also used. Because our models contain word-piece tokenisers which may split a single word into several sub-word units, we assigned each word's label to all of its sub-words and then fine-tuned our models based on the new labels. As shown in Table \ref{t:distilledNER}, DistilBioBERT and CompactBioBERT outperformed other distilled models on all the datasets. Among the continually learned models, BioDistilBERT and BioMobileBERT fared best (Table \ref{t:continualNER}), while TinyBioBERT and BioTinyBERT were the fastest and most efficient models.

For RE we used the GAD \citep{bravo2015extraction} and CHEMPROT \citep{krallinger2017overview} datasets and followed the same pre-processing used in \citet{lee2020biobert}. On the GAD dataset, we randomly selected $10\%$ of the data for the test set using a constant seed and used the rest for training. For both datasets, we trained all of our models for $3$ epochs with learning rates of $5e-5$ or $3e-5$ and a batch size of $16$. We used the latest version of CHEMPROT which has $13$ different types of relations. CompactBioBERT achieved the best results in both tasks among the distilled models (Table \ref{re:distill}), and similarly, BioDistilBERT outperformed all of our continually trained models in both tasks (Table \ref{re:continual}). 

For QA, we used the BioASQ 7b dataset \citep{tsatsaronis2015overview} and followed the same pre-processing steps as \citet{lee2020biobert}. All the models were trained with a batch size of $16$. For TinyBERT, TinyBioBERT, and BioTinyBERT a learning rate of $5e-5$ was used while for the remaining models this value was set to $3e-5$. As seen in Table \ref{qa:distil}, among our distilled models CompactBioBERT and TinyBioBERT performed best, and  among our continually learned models BioMobileBERT and BioDistilBERT outperformed other distilled models (Table \ref{qa:continual}). 

\begin{table}[ht!]
    \centering
    \caption{\label{t:distilledNER} Test results for the models that were directly distilled from the BioBERT-v1.1 on NER datasets. The $*$ symbol indicates that any direct comparison should take into account the fact that other models include over $60$M parameters, whereas TinyBioBERT has only $15$M.}
    \scalebox{0.7}{
    \begin{tabular}{L{2cm}L{2cm}P{1cm}P{2cm}P{2cm}P{2.5cm}P{2.5cm}P{2.5cm}}
        \toprule[1pt]
        Type & Task & Metrics & DistilBERT & DistilBioBERT & CompactBioBERT & TinyBioBERT$^{*}$ & BioBERT-v1.1 \\\midrule[0.5pt]
        
        Disease    & NCBI disease & P & 85.02 & 86.74 & \underline{86.91} & 82.11 & \textbf{87.23}\\
                   &  & R & 87.78 & 89.14 & \textbf{90.50} & 88.57 & \underline{90.07}\\
                   &  & F & 86.38 & 87.93 & \textbf{88.67} & 85.22 & \underline{88.62}\\

                   & BC5CDR & P & 81.57 & 84.34 & \underline{84.76} & 79.91 & \textbf{85.81}\\
                   &  & R & 82.47 & \underline{86.54} & 86.01 & 82.71 & \textbf{87.54}\\
                   &  & F & 82.01 & \underline{85.42} & 85.38 & 81.28 & \textbf{86.67}\\

        Drug/chem. & BC5CDR & P & 92.11 & \underline{94.04} & 94.03 & 91.31 & \textbf{94.47}\\
                   &  & R & 92.90 & \textbf{95.04} & 94.60 & 93.09 & \underline{95.00}\\
                   &  & F & 92.50 & \underline{94.53} & 94.31 & 92.20 & \textbf{94.73}\\

                   & BC4CHEMD & P & 90.91 & \underline{92.48} & 91.97 & 88.77 & \textbf{92.77}\\
                   &  & R & 88.19 & \underline{91.06} & 90.83 & 89.29 & \textbf{91.51}\\
                   &  & F & 89.53 & \underline{91.77} & 91.40 & 89.03 & \textbf{92.14}\\
        
        Gene/protein & BC2GM & P & 83.93 & \underline{86.11} & 85.55 & 80.49 & \textbf{87.07}\\
                     &  & R & 85.29 & 87.10 & \underline{87.90} & 84.65 & \textbf{88.17}\\
                     &  & F & 84.61 & 86.60 & \underline{86.71} & 82.52 & \textbf{87.62}\\

                     & JNLPBA & P & 73.37 & \underline{74.36} & 73.84 & 72.58 & \textbf{74.81}\\
                     &  & R & 85.90 & 86.49 & \textbf{86.98} & 86.07 & \underline{86.72}\\
                     &  & F & 79.14 & \underline{79.97} & 79.88 & 78.75 & \textbf{80.33}\\
        
        Species      & LINNAEUS & P & 83.42 & \underline{86.32} & 85.22 & 78.08 & \textbf{87.61}\\
                     &  & R & 78.21 & 80.45 & \underline{80.70} & 78.51 & \textbf{80.60}\\
                     &  & F & 80.73 & \underline{83.29} & 82.90 & 78.29 & \textbf{83.96}\\
                     
                     & Species-800 & P & 73.61 & 75.76 & \underline{76.21} & 67.89 & \textbf{76.74}\\
                     &  & R & 70.51 & 73.70 & \underline{75.19} & 71.36 & \textbf{79.03}\\
                     &  & F & 72.03 & 74.72 & \underline{75.70} & 69.59 & \textbf{77.87}\\
                   \bottomrule
    \end{tabular}}
    \vspace{10pt}
    
\end{table}

\begin{table}[ht!]
    \centering
    \caption{\label{t:continualNER} NER test results for models that were pre-trained on the PubMed dataset via the MLM objective and continual learning. Note that the models beginning with the prefix `Bio' are pre-trained, while the rest are baselines.}
    \scalebox{0.7}{
    \begin{tabular}{L{2.5cm}P{1.5cm}P{2cm}P{2cm}P{2cm}P{2cm}P{2cm}P{2.5cm}}
        \toprule[1pt]
        Task & Metrics & DistilBERT & TinyBERT & MobileBERT & BioDistilBERT & BioTinyBERT & BioMobileBERT \\\midrule[0.5pt]
        NCBI disease & P & 85.02 & 79.59 & 84.29 & \textbf{86.93} & 80.41 & \underline{86.36}\\
                     & R & 87.78 & 81.36 & \underline{88.07} & \textbf{88.31} & 85.66 & \underline{88.07}\\
                     & F & 86.38 & 80.46 & 86.14 & \textbf{87.61} & 82.95 & \underline{87.21}\\

        BC5CDR(disease) & P & 81.57 & 76.12 & 80.52 & \textbf{84.59} & 78.69 & \underline{84.03}\\
                        & R & 82.47 & 78.83 & 83.51 & \textbf{86.66} & 83.79 & \underline{85.23}\\
                        & F & 82.01 & 77.45 & 81.99 & \textbf{85.61} & 81.16 & \underline{84.62}\\

        BC5CDR(chem) & P & 92.11 & 90.19 & 92.45 & \textbf{94.70} & 89.90 & \underline{93.88}\\
                     & R & 92.90 & 86.87 & 91.95 & \textbf{94.25} & 91.83 & \underline{94.58}\\
                     & F & 92.50 & 88.50 & 92.20 & \textbf{94.48} & 90.85 & \underline{94.23}\\

        BC4CHEMD & P & 90.91 & 85.84 & 90.65 & \underline{92.18} & 88.17 & \textbf{92.29}\\
                 & R & 88.19 & 81.79 & 88.58 & \textbf{91.00} & 86.59 & \underline{90.36}\\
                 & F & 89.53 & 83.76 & 89.60 & \textbf{91.59} & 87.37 & \underline{91.31}\\
        
        BC2GM & P & 83.93 & 76.43 & 82.62 & \textbf{86.28} & 78.86 & \underline{84.44}\\
              & R & 85.29 & 77.43 & 83.09 & \textbf{87.68} & 82.36 & \underline{86.10}\\
              & F & 84.61 & 76.93 & 82.86 & \textbf{86.97} & 80.57 & \underline{85.26}\\

        JNLPBA & P & 73.37 & 71.04 & 73.18 & \underline{73.56} & 71.74 & \textbf{74.81}\\
               & R & \underline{85.90} & 83.55 & 85.54 & 85.54 & 85.14 & \textbf{86.28}\\
               & F & \underline{79.14} & 76.79 & 78.88 & 79.10 & 77.87 & \textbf{80.13}\\
        
        LINNAEUS & P & \underline{83.42} & 77.16 & 74.72 & \textbf{85.69} & 78.88 & 81.63\\
                 & R & 78.21 & 67.38 & \textbf{82.75} & 79.66 & 74.10 & \underline{82.03}\\
                 & F & 80.73 & 71.94 & 78.53 & \textbf{82.56} & 76.42 & \underline{81.83}\\
                     
        Species-800 & P & 73.61 & 66.62 & 71.76 & \textbf{74.39} & 67.80 & \underline{74.33}\\
                    & R & 70.51 & 66.04 & \textbf{77.59} & 74.98 & 73.82 & \underline{76.14}\\
                    & F & 72.03 & 66.33 & 74.56 & \underline{74.68} & 70.68 & \textbf{75.22}\\
                   \bottomrule
    \end{tabular}}
    \vspace{10pt}
    
\end{table}

        

        

\section{Discussion}
In this study, we investigated two approaches for compressing biological language models.
The first strategy was to distil a model from a biomedical teacher, and the second was to use MLM pre-training to adapt an already distilled model to a biomedical domain. Due to computational and time constraints, we trained our distilled models for $100$k steps and our continually learned models for 200k steps; as a result, directly comparing these two types of models may be unfair. We observed that distilling a compact model from a biomedical teacher increases its capacity to perform better on complex biomedical tasks while decreasing its general language understanding and reasoning. This means that while our distilled models perform exceptionally well on biomedical NER and RE (Tables \ref{t:distilledNER} and \ref{re:distill}), they perform comparatively poorly on tasks that require more general knowledge and language understanding such as biomedical QA (Table \ref{qa:distil}). 

Weaker results on QA (compared to continually learned models) suggest that by distilling a model from scratch using a biomedical teacher, the model may lose some of its ability to capture complex grammatical and semantic features while becoming more powerful in identifying and understanding biomedical correlations in a given context (as seen in Table \ref{re:distill}). On the other hand, adapting already compact models to the biomedical domain via continual learning seems to preserve general knowledge regarding natural language structure and semantics in the model (Table \ref{qa:continual}). It should be noted that the distilled models are only trained for $100$k steps and this analysis is based on the current results obtained by these models.

Furthermore, despite having nearly half as many parameters, BioMobileBERT outscored BioDistilBERT on QA. As previously stated, MobileBERT employs a unique structure that allows it to get as deep as $24$ layers while maintaining less than $30$M parameters. On the other hand, BioDistilBERT is only $6$ layers deep. Because of this architectural difference, we hypothesise that the increased number of layers in BioMobileBERT allows it to capture more complex grammatical and semantic features, resulting in superior performance in biomedical QA, which requires not only biomedical knowledge but also some general understanding about natural language. 

We trained models of varied sizes and topologies, ranging from small models with only $25$M parameters to larger models with up to 65M. In our experiments, we discovered that when fine-tuned with a high learning rate (e.g. $5e-5$), our tiny models, TinyBioBERT and BioTinyBERT, perform well on downstream tasks while our bigger models tend to perform better with a lower learning rate (e.g. $3e-5$).

In addition, we found that compact models that have been trained on the PubMed dataset for fewer training steps (e.g. $50$k) tend to achieve better results on more general biomedical datasets such as NCBI-disease which are annotated for disease mentions and concepts and perform worse on more specialised datasets like BC5CDR-disease and BC5CDR-chem which include extra domain-specific information (e.g. chemicals and chemical-disease interactions), and the reverse is true for the models that are trained longer on the PubMed dataset.

TinyBioBERT and BioTinyBERT are the most efficient models in terms of both memory and time complexity (as evidenced by Figure \ref{fig:efficieny}). DistilBioBERT, CompactBioBERT, and BioDistilBERT are the second most efficient set of models in terms of time complexity. BioMobileBERT, on the other hand, is the second most efficient model with regards to memory complexity. In conclusion, if efficiency is the most important factor, the tiny models are the most suitable resources to use. In other use cases, we recommend either the distilled models or BioMobileBERT depending on the relative importance of memory, time, and accuracy.

\label{disc}

\section{Conclusion}
\label{concl}
In this work, we employed a number of compression strategies to develop compact biomedical transformer-based models that proved competitive on a range of biomedical datasets. We introduced six different models ranging from $15$M to $65$M parameters and evaluated them on three different tasks. We found that competitive performance may be achieved by either pre-training existing compact models on biomedical data or distilling students from a biomedical teacher. The choice of distillation or pre-training is dependent on the task, since our pre-trained students outperformed their distilled counterparts in some tasks and vice versa.
We discovered, however, that distillation from a biomedical teacher is generally more efficient than pre-training when using the same number of training steps. Due to computational and time constraints, we trained all of our distilled models for $100$k steps, and for continual learning, we trained models for $200$k steps. For future work, we plan to pre-train models for $500$k to $1$M steps and publicly release the new models. In addition, since CompactBioBERT and DistilBioBERT performed similarly on most of the tasks, we plan to investigate the effect of hyperparameters on training these models in order to determine which distillation technique is more efficient. Some of the compact biomedical models proposed in this study may be used for inference on mobile devices, which we hope will open new avenues for researchers with limited computational resources.

\begin{table}[ht!]
    \centering
    \caption{\label{re:distill} Test results of the models that were directly distilled from the BioBERT-v1.1 on RE datasets. The $*$ symbol indicates that any direct comparison between TinyBioBERT and other models should account for the significance difference in model size ($15$M vs $60M$). Scores for GAD are in the binary mode and the metrics reported for CHEMPROT are macro-averaged.}
    \scalebox{0.7}{
    \begin{tabular}{L{2.5cm}L{1.5cm}P{1cm}P{2cm}P{2cm}P{2.5cm}P{2.5cm}P{2.5cm}}
        \toprule[1pt]
        Relation & Task & Metrics & DistilBERT & DistilBioBERT & CompactBioBERT & TinyBioBERT$^{*}$ & BioBERT-v1.1 \\\midrule[0.5pt]
        
        Gene–disease     & GAD & P & 77.60 & 78.76 & \textbf{80.18} & 77.20 & \underline{79.82}\\
                         &  & R & 88.15 & \underline{93.03} & 91.63 & 88.50 & \textbf{95.12}\\
                         &  & F & 82.54 & 85.30 & \underline{85.52} & 82.46 & \textbf{86.80}\\
                         
                         
        Protein–chemical & CHEMPROT & P & 47.41 & 49.90 & \textbf{52.74} & 31.02 & \underline{52.00}\\
                         &  & R & 47.89 & 50.30 & \underline{52.93} & 33.61 & \textbf{53.03}\\
                         &  & F & 47.52 & 49.79 & \textbf{52.46} & 30.33 & \underline{52.32}\\
                        \bottomrule
    \end{tabular}}
    \vspace{10pt}

\end{table}

\begin{table}[ht!]
    \centering
    \caption{\label{re:continual} Test results on RE datasets for the models that were pre-trained on PubMed  via MLM objective and continual learning. Model names beginning with the prefix `Bio' are pre-trained and the others are baselines. Scores for GAD are in the binary mode and the metrics reported for CHEMPROT are macro-averaged.}
    \scalebox{0.7}{
    \begin{tabular}{L{2cm}P{1.5cm}P{2cm}P{2cm}P{2cm}P{2cm}P{2.5cm}P{2.5cm}}
        \toprule[1pt]
        Task & Metrics & DistilBERT & TinyBERT & MobileBERT & BioDistilBERT & BioTinyBERT & BioMobileBERT \\\midrule[0.5pt]
        
        GAD & P & 77.60 & 71.42 & 76.31 & \textbf{81.36} & 74.22 & \underline{78.50}\\
            & R & 88.15 & 80.13 & 90.94 & \underline{91.28} & 83.27 & \textbf{91.63}\\
            & F & 82.54 & 75.53 & 82.98 & \textbf{86.04} & 78.48 & \underline{84.56}\\
        
        CHEMPROT & P & 47.41 & 28.50 & 47.61 & \textbf{51.56} & 31.33 & \underline{50.77}\\
                 & R & 47.89 & 27.53 & 48.67 & \textbf{51.84} & 29.56 & \underline{51.60}\\
                 & F & 47.52 & 23.18 & 47.92 & \textbf{51.48} & 25.54 & \underline{51.03}\\
                \bottomrule
    \end{tabular}}
    \vspace{10pt}

\end{table}

\begin{table}[ht!]
    \centering
    \caption{\label{qa:distil} Test results of the models that were directly distilled from the BioBERT-v1.1 on the BioASQ QA dataset. The metrics used for reporting the results are taken from the BioASQ competition and the models were assessed using the same evaluation script. The metrics are as follows: Strict Accuracy (S), Lenient Accuracy (L) and Mean Reciprocal Rank (M).}
    \scalebox{0.7}{
    \begin{tabular}{L{2cm}P{1cm}P{2.4cm}P{2.9cm}P{2.9cm}P{2.9cm}P{2.9cm}}
        \toprule[1pt]
         Task & Metrics & DistilBERT & DistilBioBERT & CompactBioBERT & TinyBioBERT$^{*}$ & BioBERT-v1.1 \\\midrule[0.5pt]
        
            BioASQ 7b & S & 20.98 & 20.98 & \underline{22.83} & 20.98 & \textbf{24.07}\\
                      & L & 29.62 & 28.39 & 29.01 & \underline{30.86} & \textbf{34.56}\\
                      & M & 24.34 & 23.79 & \underline{25.06} & 25.05 & \textbf{28.41}\\
            \bottomrule
    \end{tabular}}
    \vspace{10pt}

\end{table}

\begin{table}[ht!]
    \centering
    \caption{\label{qa:continual} BioASQ QA test results for the models that were pre-trained on the PubMed dataset via MLM objective and continual learning. The metrics used for reporting the results are taken from the BioASQ competition and the models were assessed using the same evaluation script. The metrics are as follows: Strict Accuracy (S), Lenient Accuracy (L) and Mean Reciprocal Rank (M) scores.}
    \scalebox{0.7}{
    \begin{tabular}{L{2cm}P{1cm}P{2cm}P{2cm}P{2cm}P{2.5cm}P{2.5cm}P{2.5cm}}
        \toprule[1pt]
         Task & Metrics & DistilBERT & TinyBERT & MobileBERT & BioDistilBERT & BioTinyBERT & BioMobileBERT \\\midrule[0.5pt]
        
            BioASQ 7b & S & 20.98 & 21.60 & \underline{27.77} & 25.92 & 20.37 & \textbf{29.01}\\
                      & L & 29.62 & 29.62 & \textbf{40.74} & \underline{38.88} & 32.09 & \underline{38.88}\\
                      & M & 24.34 & 24.62 & \underline{32.78} & 30.83 & 25.20 & \textbf{32.90}\\
            \bottomrule
    \end{tabular}}
    \vspace{10pt}
    
\end{table}


\section*{Funding}

This work was supported in part by the National Institute for Health Research (NIHR) Oxford Biomedical Research Centre (BRC), and in part by an InnoHK Project at the Hong Kong Centre for Cerebro-cardiovascular Health Engineering (COCHE). OR acknowledges the generous support of the Medical Research Council (grant number MR/W01761X/). DAC is an Investigator in the Pandemic Sciences Institute, University of Oxford, Oxford, UK. The views expressed are those of the authors and not necessarily those of the NHS, the NIHR, the MRC, the Department of Health, InnoHK – ITC, or the University of Oxford.

\bibliographystyle{natbib}
\bibliography{document}









\end{document}